\pdfoutput=1

\documentclass[11pt]{article}

\usepackage[]{acl}

\usepackage{times}
\usepackage{latexsym}

\usepackage[T1]{fontenc}

\usepackage[utf8]{inputenc}

\usepackage[letterspace=-26]{microtype}

\usepackage{bm}
\usepackage{booktabs}
\usepackage{stfloats}
\usepackage{float}
\usepackage{multirow}
\usepackage{algorithm}
\usepackage{algorithmic}
\usepackage{graphicx}
\usepackage{enumerate}
\usepackage{subfigure}
\usepackage{amsmath}
\usepackage{amssymb}
\usepackage{tabularx}
\usepackage{booktabs}
\usepackage{multirow}
\usepackage{tikz}

\definecolor{c0}{cmyk}{1,0.3968,0,0.2588} 
\definecolor{c1}{cmyk}{0,0.6175,0.8848,0.1490} 
\definecolor{c2}{cmyk}{0.1127,0.6690,0,0.4431} 
\definecolor{c3}{cmyk}{0.3081,0,0.7209,0.3255} 
\definecolor{c4}{cmyk}{0.6765,0.2017,0,0.0667} 
\definecolor{c5}{cmyk}{0,0.8765,0.7099,0.3647} 

\definecolor{darkgrey}{RGB}{149,149,149}
\definecolor{decentgrey}{RGB}{242,242,242}

\usetikzlibrary{calc,fit,positioning,shapes,arrows,arrows.meta,backgrounds,decorations.pathreplacing}
\pgfdeclarelayer{bg}
\pgfsetlayers{bg,main}

\newcommand{\oursFull}{Semantic-Oriented Unlabeled Priming}
\newcommand{\oursShort}{\textsc{Soup}}

\newcounter{notecounter}
\newcommand{\enotesoff}{\long\gdef\enote##1##2{}}
\newcommand{\enoteson}{\long\gdef\enote##1##2{{
			\stepcounter{notecounter}
			{\large\bf
				\hspace{1cm}\arabic{notecounter}\color{red}{ $<<<$ ##1: ##2
					$>>>$\hspace{1cm}}}}}}

\enoteson
\enotesoff

\title{\oursFull{} for Large-Scale Language Models}

\author{Yanchen Liu\footnotemark[1] \quad Timo Schick\footnotemark[2] \quad Hinrich Sch\"{u}tze\footnotemark[2]\\[0.5em]
	\footnotemark[1]\ \ Department of Informatics, Technical University of Munich, Germany \\  \footnotemark[2]\ \ Center for Information and Language Processing (CIS), LMU Munich, Germany \\[0.5em]
	{\tt yanchen.liu@tum.de}
}

\begin{document}
	\maketitle
	\begin{abstract}
		Due to the high costs associated with finetuning large language models, 
		various recent works propose to adapt them to specific tasks without any parameter updates through in-context learning.
		Unfortunately, for in-context learning there is currently no way to leverage unlabeled data, which is often much easier to obtain in large quantities than labeled examples.
		In this work, we therefore investigate ways to make use of unlabeled examples to improve the zero-shot performance of pretrained language models without any finetuning: We introduce \oursFull{} (\oursShort{}), a method that classifies examples by retrieving semantically similar unlabeled examples, assigning labels to them in a zero-shot fashion, and then using them for in-context learning. We also propose \emph{bag-of-contexts} priming, a new priming strategy that is more suitable for our setting and enables the usage of more examples than fit into the context window.
	\end{abstract}

	\section{Introduction}
	
	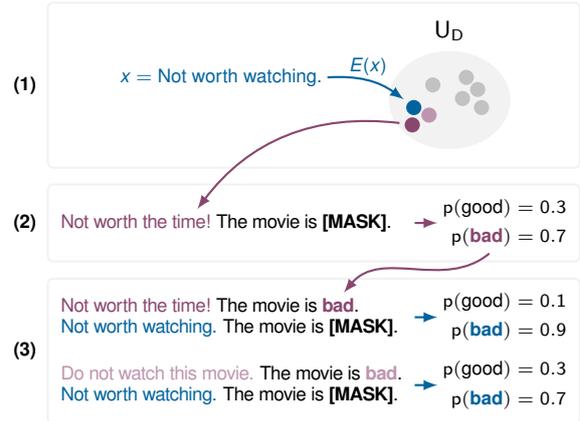
\begin{figure}
		\centering
		\tikzset{
			prompt/.style={
				font=\sffamily\scriptsize, inner sep=0, outer sep=2pt,	
			},
			box/.style={
				thick, font=\sffamily\scriptsize, align=left, draw=decentgrey, inner sep=3pt, rounded corners=2pt,
			},
			ud/.style = {
				ellipse, minimum width=0.2cm, minimum height=0.2cm, fill=decentgrey!80!black, inner sep=0, outer sep=2pt,
			},
			tarrow/.style={
				draw,->,>=latex, thick
			},
		}
		\centering
		\begin{tikzpicture}
		
		\node[](anchor){};
		\node[prompt, right=0.5cm of anchor](input){\textcolor{c0}{\emph{x} $ = $ Not worth watching.}};
		\node[below=0cm of anchor.west, anchor=west, minimum width=0.88\linewidth](input-dummy){};
		\node[right=0.8cm of input, fill=decentgrey, minimum width=1.6cm, minimum height=1.3cm, ellipse, yshift=-0.3cm](ud-circ){};
		\node[above=0cm of ud-circ](ud-headline){\footnotesize $\mathsf{U_D}$};
		\node[below=0.1cm of ud-circ, outer sep=0, inner sep=0](ud-padding){};
		
		\node[ud, right=0.25cm of ud-circ.center, yshift=-0.1cm](){};
		\node[ud, right=0.05cm of ud-circ.center, yshift=0.3cm](){};
		\node[ud, right=0.2cm of ud-circ.center, yshift=0.14cm](){};
		\node[ud, left=0.05cm of ud-circ.center, yshift=0.2cm](){};		
		\node[ud, right=0.0cm of ud-circ.center, yshift=0.02cm](){};		
		
		\node[ud, left=0.1cm of ud-circ.center, yshift=-0.2cm, fill=c2!50](n1){};				
		\node[ud, left=0.32cm of ud-circ.center, yshift=-0.33cm, fill=c2](n2){};				
		\node[ud, left=0.3cm of ud-circ.center, yshift=-0.1cm, fill=c0](e-of-x){};
		
		\path [] (input.east) edge[tarrow, bend left=20, color=c0] node [above,midway,prompt] {\emph{E}(\emph{x})} (e-of-x);
	
		\node[prompt, below=1.93cm of anchor.west, anchor=west](input-n1){\lsstyle\textcolor{c2}{Not worth the time!} The movie is \textbf{[MASK]}.\phantom{xx}};
		\node[below=0cm of input-n1.west, anchor=west, minimum width=0.88\linewidth](input-n1-dummy){};
		
		\path[] (n2) edge[tarrow, bend right=30, color=c2] ([xshift=-0.5cm]input-n1.north);
		
		\node[prompt, right=0.3cm of input-n1, align=left](scores-n1){$\begin{aligned} \mathsf{p(\text{good})} & = \mathsf{0.3} \\ \mathsf{p(\textcolor{c2}{\textbf{bad}})} & = \mathsf{0.7} \end{aligned}$};

		\path[] (input-n1) edge[tarrow, color=c2] (scores-n1);
						
		\node[prompt, below=1.25cm of input-n1.west, anchor=west, align=left](input-n1-x){\lsstyle\textcolor{c2}{Not worth the time!} The movie is \textcolor{c2}{\textbf{bad}}.\\ \lsstyle\textcolor{c0}{Not worth watching.} The movie is \textbf{[MASK]}.};
		\node[below=0cm of input-n1-x.west, anchor=west, minimum width=0.88\linewidth](input-n1-x-dummy){};
		
		\node[prompt, right=0.3cm of input-n1-x.east -| input-n1.east, align=left](scores-n1-x){$\begin{aligned} \mathsf{p(\text{good})} & = \mathsf{0.1} \\ \mathsf{p(\textcolor{c0}{\textbf{bad}})} & = \mathsf{0.9} \end{aligned}$};
	
		\node[prompt, below=0.9cm of input-n1-x.west, anchor=west, align=left](input-n2-x){\lsstyle\textcolor{c2!50}{Do not watch this movie.} The movie is \textcolor{c2!50}{\textbf{bad}}.\\ \lsstyle\textcolor{c0}{Not worth watching.} The movie is \textbf{[MASK]}.};
		\node[below=0cm of input-n2-x.west, anchor=west, minimum width=0.88\linewidth](input-n2-x-dummy){};
		
		\node[prompt, right=0.3cm of input-n2-x.east -| input-n1.east, align=left](scores-n2-x){$\begin{aligned} \mathsf{p(\text{good})} & = \mathsf{0.3} \\ \mathsf{p(\textcolor{c0}{\textbf{bad}})} & = \mathsf{0.7} \end{aligned}$};
		
		\path[] ([xshift=-0.2cm]scores-n1.south) edge[tarrow, out=220, in=50, color=c2] ([xshift=1.5cm]input-n1-x.north);
		
		%
		\path[] (input-n1-x.east -| input-n1.east) edge[tarrow, color=c0] (scores-n1-x);		
		\path[] (input-n2-x.east -| input-n1.east) edge[tarrow, color=c0] (scores-n2-x);
	
		\begin{pgfonlayer}{bg}
			\node[box, fit=(input)(ud-headline)(ud-circ)(input-dummy)(ud-padding)](box-1){};
		\end{pgfonlayer}
		
		\begin{pgfonlayer}{bg}
			\node[box, fit=(input-n1)(input-n1-dummy)(scores-n1)](box-2){};
		\end{pgfonlayer}
		
		\begin{pgfonlayer}{bg}
			\node[box, fit=(input-n1-x)(scores-n1-x)(input-n1-x-dummy)(scores-n2-x)(input-n2-x)](box-3){};
		\end{pgfonlayer}
		
		\node[left=0cm of box-1](){\scriptsize\sffamily\textbf{(1)}};
		\node[left=0cm of box-2](){\scriptsize\sffamily\textbf{(2)}};
		\node[left=0cm of box-3](){\scriptsize\sffamily\textbf{(3)}};
	
		\end{tikzpicture}
		\caption{
			Schematic representation of the steps involved in \oursShort{} for binary sentiment classification of movie reviews. (1)~\textbf{Semantic Search}: For \textcolor{c0}{a given input $x$}, we retrieve semantically similar, \textcolor{c2}{unlabeled examples} from a set $U_D$ using a sentence encoder $E$. (2)~\textbf{Self-Prediction}: We obtain zero-shot predictions for all similar examples using natural language prompts. (3)~\textbf{Bag-of-Contexts Priming}: We use the retrieved examples along with their most probable labels one at a time as in-context examples to obtain predictions for $x$; the resulting distributions over possible labels are finally averaged.
		}
	\end{figure}
	
	In recent years, there has been a trend in NLP towards larger and larger language models (LMs) \citep{radford2018improving,radford2018language,raffel2019exploring,brown2020language,fedus2021switch}. Different from prior pretrained LMs that are typically finetuned for specific downstream tasks using labeled training datasets \citep{devlin2018bert,liu2019roberta}, recent work proposes to use such large models in zero- or few-shot settings without any finetuning \citep{brown2020language,sanh2021multitask} due to the often prohibitive costs associated with training, storing and deploying large models \citep{strubell-etal-2019-energy}. In particular, \citet{brown2020language} propose \emph{priming} where training examples are simply provided as additional context together with test examples; this \emph{in-context learning} does not require updating the parameters of the model.
	
	In prior work on in-context learning, only labeled examples are used for priming \citep{brown2020language,lu2021fantastically,kumar2021reordering,min2021noisy,jiang2021know}. But in many settings, these are extremely scarce or even entirely unavailable, while unlabeled examples can easily be accessed. Unfortunately, there is currently no way to leverage unlabeled examples for priming. Other approaches for leveraging unlabeled data such as domain-adaptive pretraining \citep{gururangan-etal-2020-dont} would again require finetuning. 
	
	Therefore, we investigate how we can make use of unlabeled examples to improve the performance of large-scale language models without requiring changes to their parameters: We propose a self-supervised method called \oursFull{} (\oursShort{}), which uses unlabeled examples for in-context learning. Following the observation that semantically similar examples are better candidates as in-context examples than dissimilar ones \citep{gao2020making,DBLP:journals/corr/abs-2101-06804}, we first retrieve the semantically most similar unlabeled examples as contexts for a given input; then, we query the language model to obtain predictions for these unlabeled examples, and finally provide them along with their most likely labels as additional context. Intuitively, this approach is particularly helpful whenever the retrieved examples are easier to classify then the actual input of interest.
	
	Whereas in prior work, the in-context examples and test example are usually concatenated to form a single input that is provided to the LM, we propose to use one in-context example at a time and compute a weighted average of the so-obtained label distributions to obtain a final prediction. Besides resulting in much better performance, one benefit of this methods is that we are no longer constrained by the maximum sequence length of the used LM and thus, more neighbors can be used for priming than with the usual, concatenation-based approach. We also investigate an iterative variant of our approach where predictions for unlabeled examples are iteratively improved with \oursShort{}. On four English text classification datasets, we show that \oursShort{} improves performance of pretrained LMs.
		
	\section{Related Work}
	
	First proposed by \citet{brown2020language}, in-context learning has been studied by many recent works \citep{lu2021fantastically,kumar2021reordering,min2021noisy,jiang2021know}. Concurrent with our work, \citet{min2021noisy} also propose to perform priming with individual examples and combine the resulting predictions; however, they use a different combination technique and, similar to all prior work on in-context learning, only investigate settings with labeled examples.
	Our approach is also related to various approaches that leverage unlabeled data in few- or zero-shot settings \citep{,xie2019unsupervised,gururangan-etal-2020-dont,schick2020exploiting}, but all of them require finetuning the underlying language model.
	
	We make use of different Transformer-based sentence encoders \citep{reimers-gurevych-2019-sentence,gao2021simcse} and of textual instructions to improve model performance, an approach that was first proposed by \citet{radford2018language} and has since been investigated extensively \citep[i.a.]{schick2020exploiting,schick2021fewshot,schick2020just,gao2020making}.
	
	\section{\oursFull{}}
	
	We introduce \oursFull{} (\oursShort{}), our approach for in-context learning with unlabeled examples. To this end, let $M$ be a masked language model \citep{devlin2018bert} where for some sequence of tokens $t_1, \ldots, t_k$ that contains exactly one mask token, $M(t \mid t_1, \ldots, t_k)$ denotes the probability that $M$ assigns to $t$ at the masked position.\footnote{We focus on masked language models, but our approach can easily be transferred to autoregressive language models.} Further, let $E$ be a sentence encoder where $E(x)$ denotes the representation assigned to $x$ by $E$, and $D_U$ be a set of unlabeled examples. We consider a text classification setup where for a given input $x$, a label $y$ from a set $Y$ has to be predicted.
	
	Obtaining predictions for $x$ with \oursShort{} consists of the following steps:
	
	\begin{enumerate}
		\item \textbf{Semantic Search}: We search for unlabeled examples that are semantically most similar to $x$ using the sentence encoder $E$.
		\item \textbf{Self-Prediction}: We use $M$ to obtain predictions for these neighboring examples. 
		\item \textbf{Bag-of-Contexts Priming}: We use the neighbors and their estimated labels as additional context for priming $M$ and compute an average of the resulting label distributions to obtain a final prediction for $x$.
	\end{enumerate}
	
	\subsection{Semantic Search}
	
	Similar to prior work \citep{gao2020making,DBLP:journals/corr/abs-2101-06804}, the unlabeled examples $x_u \in D_U$ are encoded to obtain vector representations $E(x_u)$; this can be done in advance for the entire set $D_U$.
	We also compute the representation $e(x)$ of our test example and use semantic search to find the $k$ nearest neighbors of $x$ according to a specific similarity measure (e.g., cosine similarity). We denote the set of neighbors as  $N_x = \{ x_1,..., x_k \} \subseteq D_U$.  
	
	\subsection{Self-Prediction for Unlabeled Examples}
	We use $M$ to predict the label distribution for each $x_i \in N_x$, which is done similar to prior work by providing a short prompt and assigning meaningful names to all labels \citep[e.g.,][]{radford2018language,schick2020exploiting,schick2020just}. We use the same notation as \citet{schick2020exploiting,schick2020just} in that we make use of a \emph{pattern} $P$ that converts inputs $x$ into cloze questions $P(x)$ containing a single mask, and a \emph{verbalizer} $v$ that maps each label $y \in Y$ to a single token $v(y)$ representing its meaning. We define the probability of $y$ being the correct label for $x$ based on $M\left( v(y) \mid P(x) \right)$, the probability that $M$ assigns to $v(y)$ at the masked position in $P(x)$. We normalize this probability and set
	\begin{equation}
	p(y \mid x ) \propto \frac{M\left( v(y) \mid P(x) \right)}{M\left( v(y) \mid P(\varepsilon) \right)} \label{eq:normalized-prompting}
	\end{equation}
	with $\varepsilon$ denoting an empty sequence following prior work \citep{brown2020language}.

	\subsection{Priming}
	Let $\hat{N}_x = \left \{( x_i, \hat{y}_i) \right \} _{i=1}^k$ be the selected in-context neighbors with their predicted labels. Based on these semantically similar examples, we want to obtain a prediction for $x$. In the following, let $\hat{P}(x_i)$ denote $P(x_i)$ with the mask token replaced by $\hat{y}_i$.
	
	\paragraph{Concatenation Priming} Previous work usually provides all in-context examples at a time to the LM. That is, all examples are concatenated followed by the test example to obtain the input $c = [ \hat{P}(x_1), \hat{P}(x_2) ,...,\hat{P}(x_k), P(x) ]$, which is provided to the LM to get the final prediction. We refer to this variant as \textsc{Concat} priming.
	
	\paragraph{Bag-of-Contexts Priming}
	We propose \emph{bag-of-contexts} (\textsc{BoC}) priming where instead, we only use individual examples for priming and prediction each time and then compute the average of the resulting label distributions as the final prediction.
	The key advantage of this method lies in the fact that it allows us to use more examples than fit in the context window of the used model.
	
	For each in-context example $x_i \in N$, we construct a corresponding context $c_i = [ \hat{P}(x_i);P(x) ]$, similar to \textsc{Concat} with $k=1$.
	For each $c_i$, we then use the LM to obtain a distribution $q_i(y)$ over possible labels $y \in Y$ for $x$, where we employ normalization analogous to Eq.~\ref{eq:normalized-prompting}.
	Finally, we make use of a weighting function $w(x_i): N \rightarrow \mathbb{R}^+$ and compute
	\begin{equation}
	q_f(y)= \frac{1}{Z} \cdot \sum_{i=1}^{k} w(x_i) \cdot q_i(y) 
	\end{equation}
	with $Z = \sum_{i=1}^{k} w(x_i)$. We obtain the final prediction for $x$ as $\hat{y} = \mathop{\arg\max}_{y\in Y}q_f(y)$.
	We experiment with the following two weighting
	functions. \emph{uniform}: $w(x_i) = 1$.
        \emph{similarity-based}:  $w(x_i)$ is the cosine similarity between $x_i$ and $x$.
	
	\subsection{Iterative \oursShort{}}
	
	We also experiment with an iterative variant of \oursShort{} where the labels for the unlabeled examples in $D_U$ are iteratively refined. To this end, we treat each example $x_u \in D_U$ as a test example: We use \oursShort{} to reclassify $x_u$ with $D_U \setminus \{ x_u \}$ as the set of unlabeled examples.
	This means for each example $x$, we select in-context neighbors from  $D_U \setminus \{ x_u \}$ as priming contexts to allow us to refine the prediction for $x$. We can repeat this process for multiple iterations.
		
	\section{Experiments}
	
	\paragraph{Datasets} We evaluate \oursShort{} on four English datasets: IMDb \citep{maas-EtAl:2011:ACL-HLT2011} and Yelp Reviews \citep{zhang2015character} for sentiment analysis as well as AG’s News and Yahoo Questions \citep{zhang2015character} for text categorization. For each dataset, we use one of the the patterns and verbalizers introduced by \citet{schick2020exploiting}; further details can be found in Appendix~\ref{sec:dataset-details}.  For IMDb, the unlabeled in-context examples are selected from the training set of SST-2 \citep{socher2013recursive} following \citet{DBLP:journals/corr/abs-2101-06804}. For all other datasets, the in-context examples are obtained from the respective training sets.\footnote{To ensure a resource-friendly evaluation, we restrict both the unlabeled sets and the test sets to a maximum of 10,000 randomly selected examples.}

	\paragraph{Experimental Setup} For our main experiments, we use \emph{ALBERT-xlarge-v2} \citep{lan2019albert} as underlying LM and \emph{paraphrase-MiniLM-L6-v2} \citep{reimers-gurevych-2019-sentence} as sentence encoder. As the context window of ALBERT is 512 tokens, we truncate each example to 120 tokens for \textsc{Concat}. To enable a fair comparison between both priming strategies, we also set the maximum token number for \textsc{BoC} to 120. We compare \oursShort{} to zero-shot performance using only the patterns and verbalizers (``prompt only''), similar to \citet{radford2018language} and \citet{schick2021selfdiagnosis}. We do not compare to other baselines as we are not aware of other approaches that enable leveraging unlabeled data in zero-shot settings \emph{without finetuning}. For iterative \oursShort{}, we use 3 iterations to improve the labels assigned to unlabeled data.
		
	\paragraph{Results} As shown in Table~\ref{table:main-results}, when using \textsc{Concat} with $k = 3$, our method clearly performs worse than the prompt-only baseline. However, using our proposed \textsc{BoC} approach consistently outperforms not only priming with \textsc{Concat} by a large margin, but also leads to consistent improvements over our baseline on three out of four datasets for $k \geq 10$. Moreover, performance grows consistently with the number of in-context examples, with $k = 50$ resulting in improvements for each dataset considered. On average, similarity-based weighting leads to negligible gains over uniform weighting. For our iterative variant of \oursShort{}, we therefore only experiment with uniform weighting; iterative \oursShort{} leads to slight improvements for two tasks, but performs much worse than \oursShort{} for Yahoo.
	
	\begin{table}
		\setlength{\tabcolsep}{3pt}
		\footnotesize
		\centering
		\begin{tabularx}{\linewidth}{lccXXXX}
			\toprule
			& $k$ & $w(x_i)$ & \textbf{AG's}  & \textbf{Yahoo} & \textbf{IMDb} & \textbf{Yelp}  \\ \midrule
			\multicolumn{1}{l}{Prompt only}    	& --                         & --                          & 66.01          & 48.04          &        72.67       & 43.37          \\
			\multicolumn{1}{l}{\oursShort{} (\textsc{Conc.})}   & $3$                          & --                          &      43.88       &        21.96        &      54.71       &    29.56           \\
			\midrule
			\multirow{6}{*}{\oursShort{} (\textsc{BoC})} 		& \multirow{2}{*}{$3$}                          & unif.                   & 68.18          & 45.64          &         68.30      & 40.43          \\
			&                              & sim.                 & 68.18          & 45.57          &        68.31        & 40.43          \\
			& \multirow{2}{*}{$10$}                         & unif.                    & 69.64          & 49.93          &           71.03    & 44.05          \\
			&                              & sim.                 & 69.74          & 49.98          &        71.01       & 43.93          \\
			& \multirow{2}{*}{$50$}                         & unif.                    & 69.70          & \textbf{52.67}          &           72.97    & \textbf{46.21} \\
			&                              & sim.                 & \textbf{70.00} & 52.56 &       72.95        & 46.20          \\ \midrule
			\multicolumn{1}{l}{i\oursShort{} (\textsc{BoC})}		& $50$							& unif. & 69.88 & 45.22 & \textbf{73.78} & 45.79 \\
			\bottomrule
		\end{tabularx}
		\caption{Accuracy with zero-shot prompting, \oursShort{} with \textsc{Concat} and \textsc{BoC} as well as iterative \oursShort{} (i\oursShort{}) using different numbers of neighbors ($k$) and both uniform (``unif.'') and similarity-based (``sim.'') weighting.}
		\label{table:main-results}
	\end{table}
	
	\section{Analysis}
	
	We examine the influence of both increasing the language model's size and replacing the Sentence Transformer with different encoders on the performance of \oursShort{}. We also briefly discuss the efficiency of our method.
	
	\begin{table}
		\setlength{\tabcolsep}{4.5pt}
		\footnotesize
		\centering
		\begin{tabularx}{\linewidth}{lXllll}
			\toprule
			\textbf{Size} & \textbf{Method} & \textbf{AG's}  & \textbf{Yahoo} & \textbf{IMDb} & \textbf{Yelp}  \\ 
			\midrule
			xlarge & Prompt only & 66.01 & 48.04 & 72.67 & 43.37 \\		
			xlarge & \oursShort{} & 69.70 & 52.67 & 72.97 & \textbf{46.21} \\
			\midrule
			xxlarge & Prompt only & 73.51 & 57.89 & 76.67 & 45.84 \\		
			xxlarge & \oursShort{} & \textbf{74.89} & \textbf{61.82} & \textbf{79.54} & 41.00 \\
			\bottomrule
		\end{tabularx}
		\caption{Performance of a prompt-only baseline and \oursShort{} with $k = 50$ and uniform weighting using different model sizes}
		\label{table:model-sizes}
	\end{table}
	
	\paragraph{Model Size}
	
	We first focus on the impact of model size on the performance of \oursShort{}; to this end, we also evaluate our method (with $k = 50$ and uniform weighting) and the prompt-only baseline using ALBERT-xxlarge-v2 \citep{lan2019albert}, a model that is about four times as large as ALBERT-xlarge-v2. As shown in Table~\ref{table:model-sizes}, for our prompt-only baseline performance consistently improves with model size for both methods. With exception of ALBERT-xxlarge-v2 on Yelp, for which our method surprisingly leads to worse performance, \oursShort{} consistently outperforms the baseline method.
	
	\paragraph{Sentence Encoder}
	We also investigate the impact of the sentence encoder on downstream task performance. As \emph{paraphrase-MiniLM-L6-v2} was trained on a mixture of tasks that has some overlap with the tasks we evaluate on, we additionally consider \emph{msmarco-bert-base-dot-v5} \citep{reimers-gurevych-2019-sentence}, a model that was trained exclusively on MS MARCO passages \citep{bajaj2018ms}, and \emph{unsup-simcse-roberta-large} \citep{gao2021simcse}, an encoder that was trained in a fully unsupervised fashion.
	As can be seen in Table~\ref{table:sentence-encoders}, the choice of sentence encoder has little influence on performance, illustrating that performance improvements do not come from the encoder being pretrained on downstream task data.
	
	\begin{table}
		\setlength{\tabcolsep}{3pt}
		\footnotesize
		\centering
		\begin{tabularx}{\linewidth}{Xllll}
			\toprule
			\textbf{Sentence Encoder} & \textbf{AG's}  & \textbf{Yahoo} & \textbf{IMDb} & \textbf{Yelp}  \\ 
			\midrule
			paraphrase-MiniLM-L6-v2 & 69.70 & 52.67 & 72.97 & \textbf{46.21} \\
			msmarco-bert-base-dot-v5 & \textbf{69.93} & \textbf{53.04} & \textbf{74.47} & 45.82 \\		
			unsup-simcse-roberta-large & 69.76 & 52.40 & 73.90 & 45.19 \\
			\bottomrule
		\end{tabularx}
		\caption{\oursShort{} (ALBERT-xlarge-v2, $k
		= 50$, uniform weighting) is robust to
		choice of sentence encoder.}
		\label{table:sentence-encoders}
	\end{table}
		
	\paragraph{Efficiency} One disadvantage of our approach is that the number of required forward passes grows linearly with $k$. After precomputing encodings and labels for $U_D$, classifying a single example with $k=3$ took about 0.6s using a single NVIDIA GeForce GTX 1080Ti; for $k = 10$ and $k = 50$, the required times were 1.5s and 6.8s. However, performance can be improved a lot with decoder-only LMs \citep[e.g.,][]{radford2018improving,radford2018language,brown2020language}, as this enables the precomputation of contextualized representations for each $x_u \in U_D$.
		
	\section{Conclusion}
	
	We have presented \oursShort{}, a method for \emph{unlabeled priming} that classifies inputs by retrieving semantically similar unlabeled examples, classifying these examples in a zero-shot fashion and providing them as additional contexts for in-context learning. Beyond that, we have proposed a new priming strategy that leads to much better performance and scales to more than just a few examples. We have shown that with sufficiently many retrieved examples, \oursShort{} consistently leads to improved performance.
	
	\bibliography{custom}
	\bibliographystyle{acl_natbib}
	
	\clearpage
	\appendix
	
	\section{Dataset Details}
	\label{sec:dataset-details}
	
	For each task except IMDb, we use one of the patterns and verbalizers introduced by \citet{schick2020exploiting}. In the following, we describe in detail the patterns and verbalizers used.
	
	\paragraph{IMDb}
	For the IMDb Large Movie Review Dataset \citep{maas-EtAl:2011:ACL-HLT2011}, the task is to estimate the binary sentiment of a movie review based on the review’s text. We use the following pattern and verbalizer for an input review $a$:
	\begin{align*}
	P(a) & = a \text{. The movie is [MASK].} \\
	v(0) & = \text{bad} \quad v(1) = \text{good}
	\end{align*}
	
	\paragraph{Yelp}
	For the Yelp Reviews Full Star dataset \citep{zhang2015character}, the task is to estimate the rating that a customer gave to a restaurant on a 1-to 5-star scale based on their review’s text. We use the following pattern for an input text $a$:\\
	
	\noindent \centerline{$P(a) = a$. In summary, the restaurant is [MASK].}\\
	
	\noindent As a verbalizer $v$, we define:\\
	
	\begin{table}[h]
		\begin{tabular}{lll}
			$v(1) =$ terrible & $v(2)=$ bad & $v(3)=$ okay \\
			$v(4)=$ good & $v(5)=$ great 
		\end{tabular}
	\end{table} 
	
	\paragraph{AG’s News} AG’s News \citep{zhang2015character} is a task to classify a news article as belonging to one of the categories \textit{World} (1), \textit{Sports} (2), \textit{Business} (3)
	or \textit{Science/Tech} (4). We define the following pattern for an input news text $a$:\\
	
	\noindent \centerline{$P(a) = a$. News Category: [MASK].}\\
	
	\noindent Intuitively, we use a verbalizer that maps 1–4 to ``World'', ``Sports'', ``Business'' and ``Science'', respectively.
	
	\paragraph{Yahoo}
	Yahoo Questions \citep{zhang2015character} is a text classification dataset. Given a question and an answer, the text has to be classified to one of ten possible categories. We make use of the following pattern for a input question $a$ and an answer $b$:\\ 
	
	\noindent \centerline{$P(a,b) = a~b$. Question Category: [MASK].}\\
	
	\noindent Our verbalizer maps labels 1–10 to the tokens ``Society'', ``Science'', ``Health'', ``Education'', ``Computer'', ``Sports'', ``Business'', ``Entertainment'', ``Relationship'' and ``Politics''.

\end{document}